\documentclass[journal]{IEEEtran}

\usepackage{algorithm}
\usepackage{algorithmicx}
\usepackage[noend]{algpseudocode} 

\usepackage[colorlinks=true, allcolors=blue]{hyperref}

\usepackage{cite}
\usepackage{acro}
\usepackage{graphicx}
\usepackage{float}
\usepackage{xcolor}
\usepackage{amsmath}
\usepackage{amssymb}
\usepackage{array}
\usepackage{url}

\newcommand{\Fig}{Fig.}

\newcommand{\Tab}{Table}

\DeclareAcronym{ROI}{
short=ROI,
long=region of interest,
}

\DeclareAcronym{IOU}{
short=IOU,
long=
,
}

\DeclareAcronym{cIOU}{
short=cIOU,
long=circle intersection over union,
}

\DeclareAcronym{DoF}{
short=DoF,
long=degrees of freedom,
}

\DeclareAcronym{CPL}{
short=CPL,
long=Center Point Localization,
}

\begin{document}
%
%
%
%

\title{Circle Representation for Medical Instance Object Segmentation}
\author{Juming Xiong, Ethan H. Nguyen, Yilin Liu, Ruining Deng, Regina N Tyree, Hernan Correa, Girish Hiremath, Yaohong Wang, Haichun Yang, Agnes B. Fogo, and Yuankai Huo, \IEEEmembership{Senior Member, IEEE}

\thanks{This work was supported in part by NIH R01DK135597 (Huo), Vanderbilt University Seeding Success Grant (Huo), NIH NIDDK DK56942(ABF), and DoD
HT9425-23-1-0003(HCY) for support. \emph{(Corresponding author: Yuankai Huo. Email: yuankai.huo@vanderbilt.edu)}}

\thanks{J. Xiong, E. H. Nguyen, Y. Liu, R. Deng, Y. Huo were with the Department of Electrical Engineering and Computer Science, Vanderbilt University, Nashville, TN 37235 USA.}

\thanks{R. N. Tyree, H. Correa, Y. Wang, H. Yang, J. T. Roland, A. B. Fogo were with the Department
of Pathology, Vanderbilt University Medical Center, Nashville,
TN, 37215, USA.}

\thanks{G. Hiremath was with the Department
of Pediatrics, Vanderbilt University Medical Center, Nashville,
TN, 37215, USA.}

}

\maketitle

\makeatletter
\def\ps@IEEEtitlepagestyle{%
  \def\@oddhead{%
    \parbox{\textwidth}{\centering\scriptsize
      License: CC-BY 4.0 \par
      Comment: Accepted for publication at the Journal of Machine Learning for Biomedical Imaging (MELBA) \url{https://melba-journal.org/2025:024} \par
      Journal-ref: Machine.Learning.for.Biomedical.Imaging. 3 (2025) \par
      DOI: \href{https://doi.org/10.59275/j.melba.2025-8bad}{10.59275/j.melba.2025-8bad}%
    }%
  }%
  \def\@evenhead{\@oddhead}%
  \def\@oddfoot{}%
  \def\@evenfoot{}%
}
\makeatother

\maketitle

\begin{abstract}
Recently, circle representation has been introduced for medical imaging, designed specifically to enhance the detection of instance objects that are spherically shaped (e.g., cells, glomeruli, and nuclei). Given its outstanding effectiveness in instance detection, it is compelling to consider the application of circle representation for segmenting instance medical objects. In this study, we introduce CircleSnake, a simple end-to-end segmentation approach that utilizes circle contour deformation for segmenting ball-shaped medical objects at the instance level. The innovation of CircleSnake lies in these three areas: (1) It substitutes the complex bounding box-to-octagon contour transformation with a more consistent and rotation-invariant bounding circle-to-circle contour adaptation. This adaptation specifically targets ball-shaped medical objects. (2) The circle representation employed in CircleSnake significantly reduces the degrees of freedom to two, compared to eight in the octagon representation. This reduction enhances both the robustness of the segmentation performance and the rotational consistency of the method. (3) CircleSnake is the first end-to-end deep instance segmentation pipeline to incorporate circle representation, encompassing consistent circle detection, circle contour proposal, and circular convolution in a unified framework. This integration is achieved through the novel application of circular convolution within the context of circle detection and instance segmentation. In practical applications, such as the detection of glomeruli, nuclei, and eosinophils in pathological images, CircleSnake has demonstrated superior performance and greater rotation invariance when compared to benchmarks. The code has been made publicly available at: ~\url{https://github.com/hrlblab/CircleSnake}.
\end{abstract}


\begin{figure*}[t]
\begin{center}
\includegraphics[width=1\textwidth]{{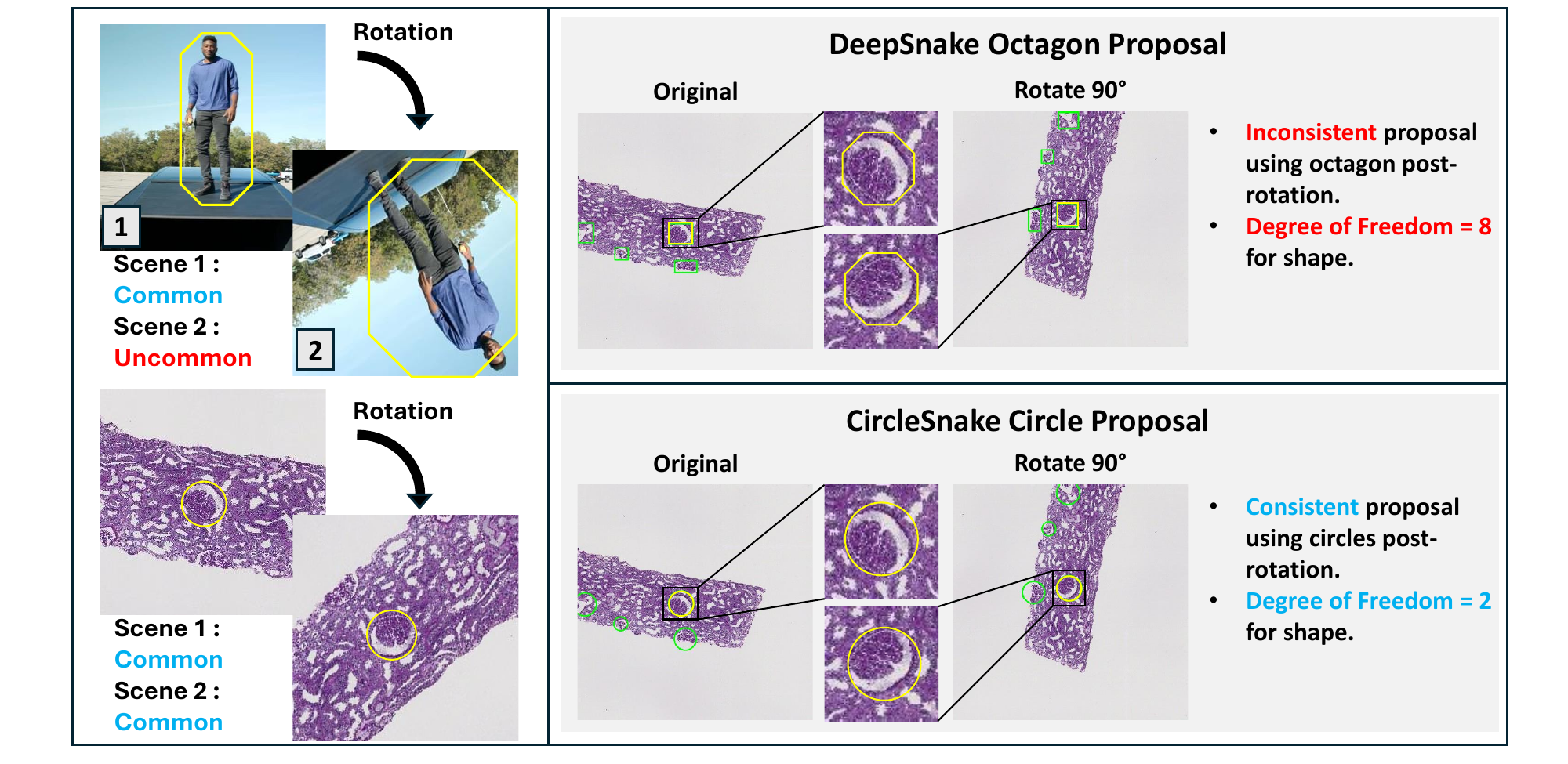}}
\end{center}
\caption{\textbf{Comparison of polygon representation and the proposed circle representation.} The left panel
shows that samples of glomeruli can be scanned at any angle of rotation. The right panel highlights the difference between the octagon proposal and the proposed circle proposal on ball-shaped objects. The proposed CircleSnake yields a more rotation consistent representation while using fewer DoF.} 
\label{Fig.1} 
\end{figure*}

\section{Introduction}

\IEEEPARstart {T}{he} \emph{deep} learning model for natural images is frequently applied in the context of medical imaging to identify and highlight regions of interest within medical images~\cite{lo2018glomerulus, LIANG2022103276, xiong2023deep}. However, medical images often present uniq-ue challenges compared to natural images. Unlike natural images which are typically displayed in a fixed orientation, medical images can be viewed at various angles~\cite{yang2020circlenet, nguyen2021circle}, making traditional rectangular bounding boxes less consistent across different angles (\Fig~\ref{Fig.1}). In such cases, circles may offer more flexible and consistent representations of delineating objects or areas of interest. This representation allows for a more consistent and orientation-independent identification of objects, which is crucial in biomedical tissue quantification across different acquisition scenarios. Moreover, the detection method can yield less rotational variances often seen in medical imagery, ensuring that key features are accurately captured regardless of which angle the image is acquired, especially when detecting ball-shaped objects~\cite{nguyen2022circlesnake}. 

Recently, the concept of circle representation has been introduced as a representation optimized for medical imaging, particularly for ball-shaped object detection, such as glomeruli, nuclei, inflammatory cells such as eosinophils, and tumors~\cite{nguyen2022circlesnake,yang2020circlenet, nguyen2021circle,luo2021scpmnet,liu2023eosinophils,zhang2023circleformer}. The exceptional effectiveness of circle representation makes it an attractive candidate for adaptation in the field of instance object segmentation.

In this paper, we propose a simple contour-based end-to-end instance segmentation method that utilizes the circle representation, called CircleSnake, for the robust segmentation of ball-shaped medical objects. The ``bounding circle" is introduced for both detection and initial contour representation on the ball-shaped objects. Once the center location of the lesion is obtained, only degrees of freedom (DoF) = 2 is required to form the bounding circle, while DoF = 8 is required for the bounding octagon. Briefly, the contribution of this study is threefold:

$\bullet$ \textbf{Circle Representation}: Our proposal introduces a unified circle representation pipeline for the segmentation of ball-shaped biomedical objects. This pipeline includes three integrated components: (1) circle detection, (2) circle contour proposal, and (3) circular convolution. It achie-ves superior segmentation performance while requiring a reduced DoF for fitting.

$\bullet$ \textbf{Optimized Biomedical Object Segmentation}:  To the best of our knowledge, CircleSnake, our proposed meth-od, represents the first instance of a contour-based end-to-end segmentation approach that is optimized for ball-shaped biomedical objects.

$\bullet$ \textbf{Rotation Consistency}: The proposed circle representation results in less DoF of fitting, improved segmentation efficiency, and enhanced rotation consistency. As shown in (\Fig~\ref{Fig.1}), tissue samples can be viewed at various angles. Consequently, improved rotational consistency could enhance the robustness in detecting identical objects within the same tissue, potentially leading to increased reproducibility in image analysis.

This study expands upon our prior conference paper
\newline \cite {nguyen2022circlesnake} by introducing fundamental enhancements in this manuscript. (1) More Comprehensive Experiments: We conduct more rigorous and comprehensive experiments using three public and in-house datasets, featuring various ball-shaped objects such as glomeruli, nuclei, and cells. (2) Methodological Depth: The methodology is presented with greater depth, including additional benchmarks, comprehensive mathematical derivations, and an elaborate experimental design. (3) Rotation Invariance Analysis: We provide deeper analyses of rotation invariance to better evaluate performance under varying conditions. (4) Reproducible Open-source Research: The open-source implementation has been updated to support both instance object detection and instance segmentation in a unified codebase \url{https://github.com/hrlblab/CircleSnake}.

\section{Related Works}
\subsection{Instance Segmentation}
Mask-based instance segmentation can be further divided into two-stage~\cite{he2018mask,8917599} and one-stage methods~\cite{bolya2019yolact, ying2019embedmask}. 

Two-stage methods usually perform detection in two steps: $(i)$ region proposal and $(ii)$ object classification and segmentation mask regression. Mask-RCNN~\cite{he2018mask} establishes the foundation for two-stage, binary mask based detection systems. Mask-RCNN~\cite{he2018mask} consists of a region proposal network (RPN) and a Region-based CNN (R-CNN) approach ~\cite{girshick2014rich, girshick2015fast} that is capable of pixel-level object instance segmentation. To achieve pixel-level precision, these binary masks classify pixels as part of the object (foreground) or not (background), simplifying the computational process and allowing for fast and accurate delineation of object boundaries. 
Following the introduction of Mask-RCNN, numerous algorithms have been developed to enhance its effectiveness, including different architectures~\cite{8578520, chen2019hybrid}, different backbone~\cite{s23083853}, transfer learning strategy~\cite{8950115}, different training strategy~\cite{e23091160}, and feature aggregation mechanisms~\cite{liu2018path}. While these two-stage detectors can achieve state-of-the-art results, they are often structurally complex and need more time to infer.

One-stage methods eliminate the region proposal step and encapsulate all computations in a single network. With the introduction of YOLACT~\cite{liu2016ssd}, these types of methods have attracted academic attention for their high computational efficiency. YOLACT employs separate prediction heads that simultaneously identify object classes, bounding boxes, and mask coefficients for each detected object. The final instance masks are dynamically assembled by combining these prototype masks with the specific mask coefficients. After YOLACT, There were many methods to improve its performance by using different feature wrapping module~\cite{liu2021yolactedge}, multi-scale feature extraction~\cite{ZENG20229419}, and adjusting model structure~\cite{Bolya_2022}. Now, the performance of one-stage object detectors that use binary masks is nearly equivalent to that of two-stage detectors with anchors, yet they offer quicker inference times.

\subsection{Medical Object Segmentation}

\subsubsection{Binary Mask-Based Methods}

In the existing literature, several works have focused on pixel-based instance \textcolor{blue}{segmentation} methodologies within regional proposals at the pixel level. Notable examples include the works of ~\cite{gadermayr2017cnn,bueno2020glomerulosclerosis,kannan2019segmentation,ginley2019computational}, which tackle the challenges of feature extraction and precise pixel-level prediction through varied strategies. For instance, ~\cite{gadermayr2017cnn} refines convolutional architectures to capture intricate details, while ~\cite{bueno2020glomerulosclerosis, kannan2019segmentation, ginley2019computational} address segmentation challenges in specific pathologies, such as glomeruli, using specialized convolutional networks. Moreover, the Mask R-CNN framework introduced by ~\cite{he2018mask} represents a significant advancement by unifying detection and segmentation into a single model. This integration not only underscores its impact on instance segmentation but also reinforces its relevance in the broader context of medical image analysis. Despite these promising developments, the structural complexity and high computational cost of these methods continue to pose challenges for real-time applications.

\subsubsection{Contour-based Methods}

\begin{figure}[h]
\begin{center}
\includegraphics[width=0.48\textwidth]{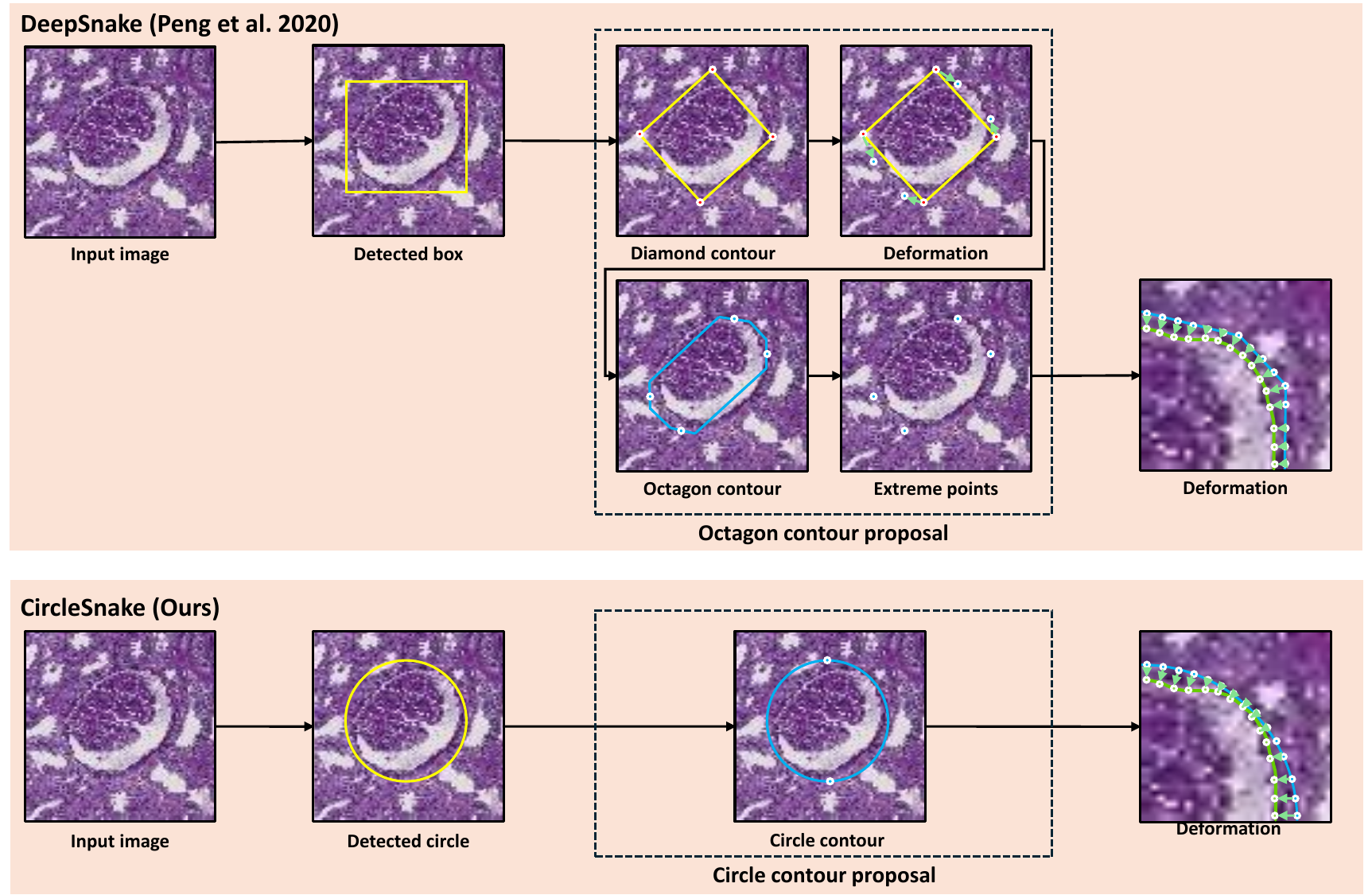}
\end{center}
   \caption{\textbf{Circle contour proposal.} This figure presents the differences between the ``bounding box to octagon contour" representation and the proposed ``bounding circle to circle contour" representation. Our circle contour proposal avoids relying on complex extreme points and deformation-based contour generation by introducing a straightforward circle proposal. This approach seamlessly connects circle detection and deformation-based segmentation in an end-to-end manner without incurring additional computational overhead. In other words, the circle detection step itself can be directly employed as a circle contour proposal.}
\label{Fig.2}
\end{figure}

Contour-based methods such as DeepSnake~\cite{peng2020deep} have the potential to be faster and simpler.~\cite{Huang2022AUB} adapts ideas from  U-Net~\cite{ronneberger2015unet} and DeepSnake~\cite{peng2020deep} to right ventricular segmentation. However, this methodology exhibits suboptimal performance when applied to circular objects in pathology images, and the computational simplicity is not adequately achieved (\Fig~\ref{Fig.2}).

\begin{figure*}[t]
\begin{center}
\includegraphics[width=1\textwidth]{{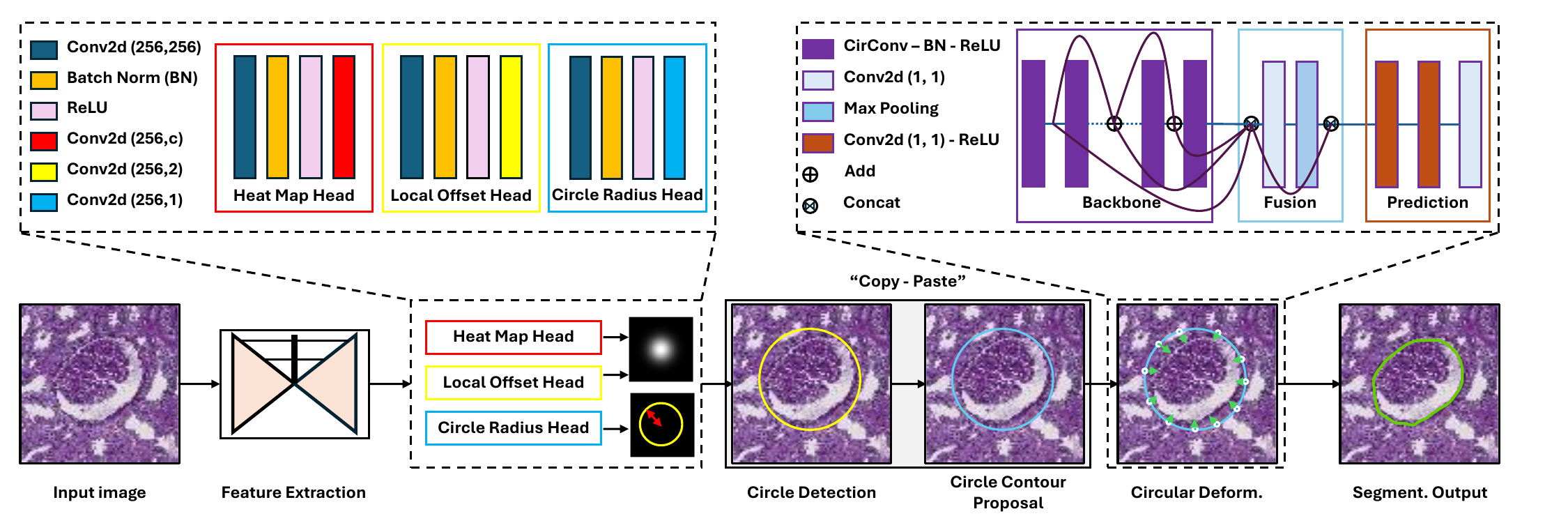}}
\end{center}
\caption{\textbf{Overview of CircleSnake.}  A backbone network functions as a feature extractor for the subsequent three head networks. The head networks for heatmap and local offset identify the circle's center point, whereas the head network for circle radius calculates the circle's radius to achieve bounding circle detection. This bounding circle establishes the initial contour. Subsequently, this contour deforms to the final contour through the use of a circular convolutional network} 
\label{Fig.3} 
\end{figure*}

\section{Methods}

The proposed CircleSnake method, shown in \Fig~\ref{Fig.3}, is an instance segmentation method inspired by the DeepSnake method~\cite{peng2020deep}. DeepSnake is a deep learning-based method for object contour detection and instance segmentation that integrates active contour models, known as "snakes," with convolutional neural networks to accurately delineate object boundaries by iteratively refining polygonal contours initialized from bounding boxes. CircleSnake, as described, modifies this approach by replacing bounding boxes with bounding circles for the initial contour proposals, utilizing circle detection algorithms to directly generate contours that are better suited for circular objects. The primary difference between DeepSnake and CircleSnake lies in their initialization strategies: while DeepSnake uses bounding boxes to create initial polygonal contours suitable for various object shapes, CircleSnake employs bounding circles to provide a closer initial approximation for circular objects, potentially enhancing efficiency and accuracy by reducing the number of refinement iterations needed. Both methods use neural networks to adjust and refine the contours, but CircleSnake's approach may offer advantages in applications involving predominantly circular objects by offering a better starting point for contour refinement compared to the more general-purpose initialization used in DeepSnake. CircleSnake aggregates the circle detection~\cite{yang2020circlenet, nguyen2021circle, nguyen2022circlesnake} and circle contour proposals for initial object segmentation (\Fig~\ref{Fig.1}). This method involves starting with a basic circular contour, which is then refined through a process of circular convolution. This process adjusts the vertices of the contour to align more closely with the actual boundary of the object (\Fig~\ref{Fig.2}). This adjustment is repeated in several iterations to accurately capture the final shape of the object for segmentation purposes. In contrast to the DeepSnake method, which depends on bounding boxes and polygonal contours, the proposed strategy is more straightforward and consistent.

\subsection{Circle Object Detection}

The method for detecting circular objects in this approach is based on the CircleNet framework~\cite{nguyen2022circlesnake}. The $I \in R^{W \times H \times 3}$ is an input image with height $H$ and width $W$. The Center Point Localization network produces the center localization of each object within a heatmap $\hat Y \in [0,1]^{\frac{W}{R} \times \frac{H}{R} \times C}$ where $C$ is the number of candidate \textcolor{blue}{classes} and $R$ is a downsampling factor. Within the heatmap, $\hat Y_{xyc} = 1$ is the center of the lesion and $\hat Y_{xyc} = 0$ is the background. All these terminologies are defined by~\cite{zhou2019objects}. Following convention, ~\cite{law2018cornernet, zhou2019objects}, the target center point is splat on a heatmap as a 2D Gaussian kernel:
\begin{equation}
{Y_{xyc} = \exp\left(-\frac{(x-\tilde p_x)^2+(y-\tilde p_y)^2}{2\sigma_p^2}\right)}
\end{equation}
\noindent where the $x$ and $y$ are the center point of the ground truth, $\tilde p_x$ and $\tilde p_y$ are the downsampled ground truth center point, and $\sigma_p$ is the kernel standard deviation. The training loss is $L_{k}$ penalty-reduced pixel-wise logistic regression with focal loss~\cite{lin2017focal}:

\begin{equation}
L_k = \frac{-1}{N} \sum_{xyc}
    \begin{cases}
    (1 - \hat{Y}_{xyc})^{\alpha} 
    \log(\hat{Y}_{xyc}) & \!\text{if}\ Y_{xyc}=1\\
    \begin{array}{c}
    (1-Y_{xyc})^{\beta} 
        (\hat{Y}_{xyc})^{\alpha}\\
        \log(1-\hat{Y}_{xyc})
        \end{array}
        & \!\text{otherwise}
    \end{cases}
\end{equation}
        
\noindent where the hyper-parameters $\alpha$ and $\beta$ are set 2 and 4 to kept the same as~\cite{lin2017focal}. To further refine the prediction location, the $\ell_1$-norm offset prediction loss $L_{off}$ is used.

To ascertain the central point, we propose identifying the top $n$ peaks, where $n$ is set to 100. Each peak's value is either greater than or equal to its 8-connected neighbors. These $n$ selected center points are denoted as $\hat{\mathcal{P}} = {(\hat x_i, \hat y_i)}_{i = 1}^{n}$. For each object, its center point is represented by an integer coordinate $(x_i,y_i)$, which is derived from $\hat Y{x_iy_ic}$ and $L_{k}$. Additionally, the offset $(\delta \hat x_i, \delta \hat y_i)$ is calculated from $L_{off}$. The formulation of the bounding circle incorporates both the center point $\hat{p}$ and the radius $\hat{r}$, thereby creating a comprehensive representation of the spatial domain:
    
\begin{equation}
 \hat{p} = (\hat x_i + \delta \hat x_i ,\ \ \hat y_i + \delta \hat y_i). \quad \hat{r} = \hat R_{\hat x_i,\hat y_i}.
\end{equation}
where $\hat R  \in \mathcal{R}^{\frac{W}{R} \times \frac{H}{R} \times 1}$ contains the radius prediction for each pixel, optimized by 
\begin{equation}
    L_{radius} = \frac{1}{N}\sum_{k=1}^{N} \left|\hat R_{p_k} - r_k\right|.
\end{equation}
where $r_k$ is the ground truth radius for each object $k$. Finally, the overall objective is
\begin{equation}
    L_{det} = L_{k} + \lambda_{radius} L_{radius} + \lambda_{off}L_{off}.
\label{eq:total_loss}
\end{equation}
Following~\cite{zhou2019objects}, we set $\lambda_{radius} = 0.1$ and $ \lambda_{off} = 1$.

\subsection{Circle Contour Proposal and Deformation}

Bounding circle-based detection for each target object is accomplished using CircleNet-based object detection, as described in~\cite{nguyen2021circle}. Directly derived from the circle representation shown in \Fig~\ref{Fig.2}, the initial circle contour proposal marks a departure from the previously complex deformation process and the use of extreme point-based octagon contour proposals. This octagon is formed by four extreme points in DeepSnake: the topmost, leftmost, bottommost, and rightmost pixels of an object, respectively. This innovation simplifies and standardizes the contour proposal approach. Defined by its radius and center point, the circle proposal is established by uniformly sampling $N$ initial points \newline ${\mathbf{x}^{circle}i | i = 1, 2,..., N}$ from the circle contour, beginning at the top-most point $x_{1}^{circle}$. Correspondingly, the ground truth contour is created by clockwise sampling of $N$ vertices along the boundary of the object. Adhering to the guidelines in~\cite{peng2020deep}, the value of $N$ is fixed at 128, facilitating a consistent and streamlined process.

For a contour with $N$ vertices denoted as \newline ${\mathbf{x}^{circle}_i | i = 1, ..., N}$, we initiate by creating feature vectors for each vertex. The feature maps are extracted by applying a CNN backbone to the input image. This CNN backbone is shared with the detector in our instance segmentation pipeline. The image feature is computed using bilinear interpolation at the vertex coordinate. The feature $f^{circle}_i$ corresponding to a vertex $\mathbf{x}^{circle}_i$ comprises a combination of the learning-based features and the vertex's coordinates, represented as \newline $[F(\mathbf{x}^{circle}_i); \mathbf{x}^{circle}_i]$. $F$ signifies the feature maps. These input features are then treated as a one-dimensional discrete signal  $f: \mathbb{Z} \to \mathbb{R}^D$ on the contour of the circle. In line with the approach detailed in~\cite{peng2020deep}, circular convolution is employed for the learning of features. The circular convolution is defined as:

\begin{equation}
    (f_N^{circle} \ast k)_i = \sum_{j = -r}^r (f_N^{circle})_{i + j}k_j,
\end{equation}

\noindent where $k: [-r, r] \to \mathbb{R}^D$ is a learnable kernel function, while the operator $\ast$ is the standard convolution. Following~\cite{peng2020deep}, the kernel size of the circular convolution is fixed to be nine.

The CircleSnake model adopts the architecture of DeepSnake~\cite{peng2020deep} for its convolutional implementation, structured into three main components: backbone, fusion, and prediction. The backbone consists of eight layers, each composed of "Conv-Bn-ReLU" sequences. This segment incorporates residual skip connections and utilizes standard convolutional layers for feature extraction. The fusion block is designed to integrate information across contour points at multiple scales. In this process, features from all layers in the backbone are combined and passed through a 1×1 convolutional layer followed by a max pooling layer. The prediction head, serving as the final component of the network, is equipped with three 1×1 convolutional layers. Its primary function is to output vertex-wise offsets. Regarding the loss function, it is specifically designed to accommodate the iterative contour deformation process within the CircleSnake framework:


\begin{equation}
    L_{iter} = \frac{1}{N}\sum_{i=1}^{N} l_{1}(\tilde x_{i}^{circle} - x_{i}^{gt}).
\end{equation}
where $x_{i}^{gt}$ is the ground truth boundary point and $\tilde x_{i}^{circle}$ is the deformed contour point. Following~\cite{peng2020deep}, we regress the $N$ offsets in 3 iterations. 

\section{Experimental Design}
\subsection{Data}
\subsubsection{Glomeruli Dataset}
To construct a dataset for glomeruli analysis, renal biopsy whole slide images (WSIs) were obtained and annotated. Kidney tissues underwent routine processing, embedding in paraffin, and were sectioned to a thickness of 3 $\mu m$, followed by staining with hematoxylin and eosin (HE), periodic acid–Schiff (PAS), or Jones. The samples were anony-mized, with studies receiving Institutional Review Board (IRB) approval by Vanderbilt University Medical Center (VUMC) under protocol number 202459 on December 28, 2020. The dataset comprised 704 glomeruli from 42 biopsies for training, 98 from 7 for validation, and 147 from 7 for testing. Given the relative size of a glomerulus~\cite{puelles2011glomerular}, the original high-resolution images (0.25 $\mu m$/ pixel) were downsampled to 4$\mu m$/pixel. Image patches of 512$\times$512 pixels containing at least one glomerulus were randomly sampled, resulting in a comprehensive dataset with 7,040 training, 980 validation, and 1,470 testing images.

\subsubsection{Nuclei Dataset}
The 2018 Multi-Organ Nuclei Segmentation challenge data-set includes 30 tissue images, each measuring 1000$\times$1000 pixels, and contains 21,623 manually annotated nuclear boundaries. These images we-re extracted from various organs sourced from \emph{The Cancer Genomic Atlas} (TCGA) using H\&E-stained WSIs at 40× magnification. Additionally, a separate testing dataset with 14 images of the same dimensions was created, including lung and brain tissues exclusively for testing. To create training and validation datasets, we randomly selected 10 patches sized at 512$\times$512 pixels from each of the 30 training/validation images, resulting in a total of 300 images. Similarly, 140 images were obtained from the 14 testing images. This dataset comprised 200 training, 100 validation, and 140 testing images. It's worth noting that the 2018 Multi-Organ Nuclei Segmentation challenge dataset contains more objects (21,623 nuclei) than the glomeruli dataset (802 glomeruli) before any data augmentation. The 2018 Multi-Organ Nuclei Segmentation challenge dataset is publicly accessible for research purposes.

\subsubsection{Eosinophils Dataset}
All image patches were cut from 50 WSIs from the Pediatric Eosinophils dataset with a 40$\times$ objective lens\cite{liu2023eosinophils}. The Eosinophils dataset contains over 12,000 annotations in 50 annotated WSIs. All human subjects research got IRB approval by VUMC under protocol number 230223 on March 2, 2023. To enhance the model's discernment of eosinophils characteristics, we established four distinct classes which include eosinophils (eos), papilla eos, red blood cells (RBC), and RBC clusters. To evaluate the performance of models, the dataset was divided into three subsets: train, val (validation), and test, with a split ratio of 7:1:2. Specifically, there are 4,842 annotations for “eos”, 2,789
annotations for “papillae eos”, 426 annotations for “RBC”, and 524 annotations for “RBC Cluster” in the training dataset. 690 annotations for “eos”, 430 annotations for “papillae eos”, 40 annotations for “RBC”, and 12 annotations for “RBC Cluster” are in the validation dataset. 1,389 annotations for “eos”, 813 annotations for “papillae eos”, 100 annotations for “RBC”, and 151 annotations for “RBC Cluster” are in the validation dataset.

\subsection{Experimental Design}
The implementation of CircleSnake's object instance segmentation and backbone networks followed DeepSnake's official PyTorch implementations. The Common Objects in Context (COCO) pre-trained model~\cite{lin2014microsoft} was used to initialize all models. All experiments were conducted on the same workstation with a 24 GB Nvidia RTX A5000. For both the glomeruli and Eos experiments, the hyperparameters were set to maximum epoch = 50, learning rate = $5e-4$, batch size = 16, and optimizer = Adam. For the nuclei experiment, the batch size was set to 4 due to memory constraints.

As baseline methods, Faster-RCNN~\cite{ren2015faster}, Mask-RCNN~\cite{he2018mask}, CenterNet~\cite{zhou2019objects}, DeepSnake~\cite{peng2020deep},  were chosen for their superior object detection and object instance segmentation performance. ResNet-50~\cite{he2016deep} and deep layer aggregation (DLA) network~\cite{yu2018deep} were used as backbone networks for these different methods. For CircleSnake, we followed the original implementation~\cite{nguyen2021circle} and used DLA for the backbone networks. 

To ensure the experiment's fairness, no data augmentation was performed on any model during training, and no hyperparameter optimization was performed for any specific model. During validation, all models were tested with the best epoch.

\subsection{Evaluation Metrics}
In the domain of computer vision, the COCO evaluation metrics have gained prominence as a means to assess the performance of object detection and segmentation algorithms. These metrics were developed as an integral component of the COCO dataset and challenge, offering a standardized framework for quantifying the accuracy and efficacy of such algorithms. In the evaluation process, we employed several key metrics, including Average Precision (AP), $AP_{50}$ (indicating Intersection over Union, or IOU, threshold at 0.5), $AP_{75}$ (IOU threshold at 0.75), $AP_{S}$ (pertaining to objects at a small scale with area less than $32^2$), and $AP_{M}$ (associated with objects at a medium scale with area greater than $32^2$ and less than $96^2$). In the context of segmentation, the Dice score was also utilized as part of the evaluation criteria. During the evaluation process, all evaluations were conducted using consistent threshold settings.

\subsection{Rotation Consistency Score}
To rigorously evaluate the resilience and precision of our model across varying angles, we have adopted the Rotation Consistency Score as a key metric~\cite{nguyen2021circle}. The rotation consistency score is calculated by using average Dice score (before and after rotation), where 1 means all boxes/circles overlapped while 0 means no boxes/circles overlapped. A higher rotation consistency score indicates that the predicted rotated test image, when returned to its original orientation, aligns more accurately with the unrotated test image. This results in a greater overlap during visualization, demonstrating better consistency in the model's predictions. This decision is rooted in the necessity to rotate the original test images by 90 degrees, a choice specifically made to mitigate the effects of intensity interpolation that might arise with arbitrary angles. By employing this structured rotation, we can effectively transform the segmentation masks from the rotated images back to their original orientation. This allows for direct comparison and calculation of the Rotation Consistency Score against the manually segmented results. We conducted two phases of experiments, where the first rotation process was only implemented during the test phase and was intentionally not used as a data augmentation strategy during the training of all methods. The second rotation process was also implemented during the test phase, but 50\% of the data was rotated during the training of all methods. This approach ensures a focused assessment of the model's performance in handling rotational variations, providing a clear insight into its robustness and accuracy under controlled conditions.

\begin{figure*}
\begin{center}
\includegraphics[width=1\textwidth]{{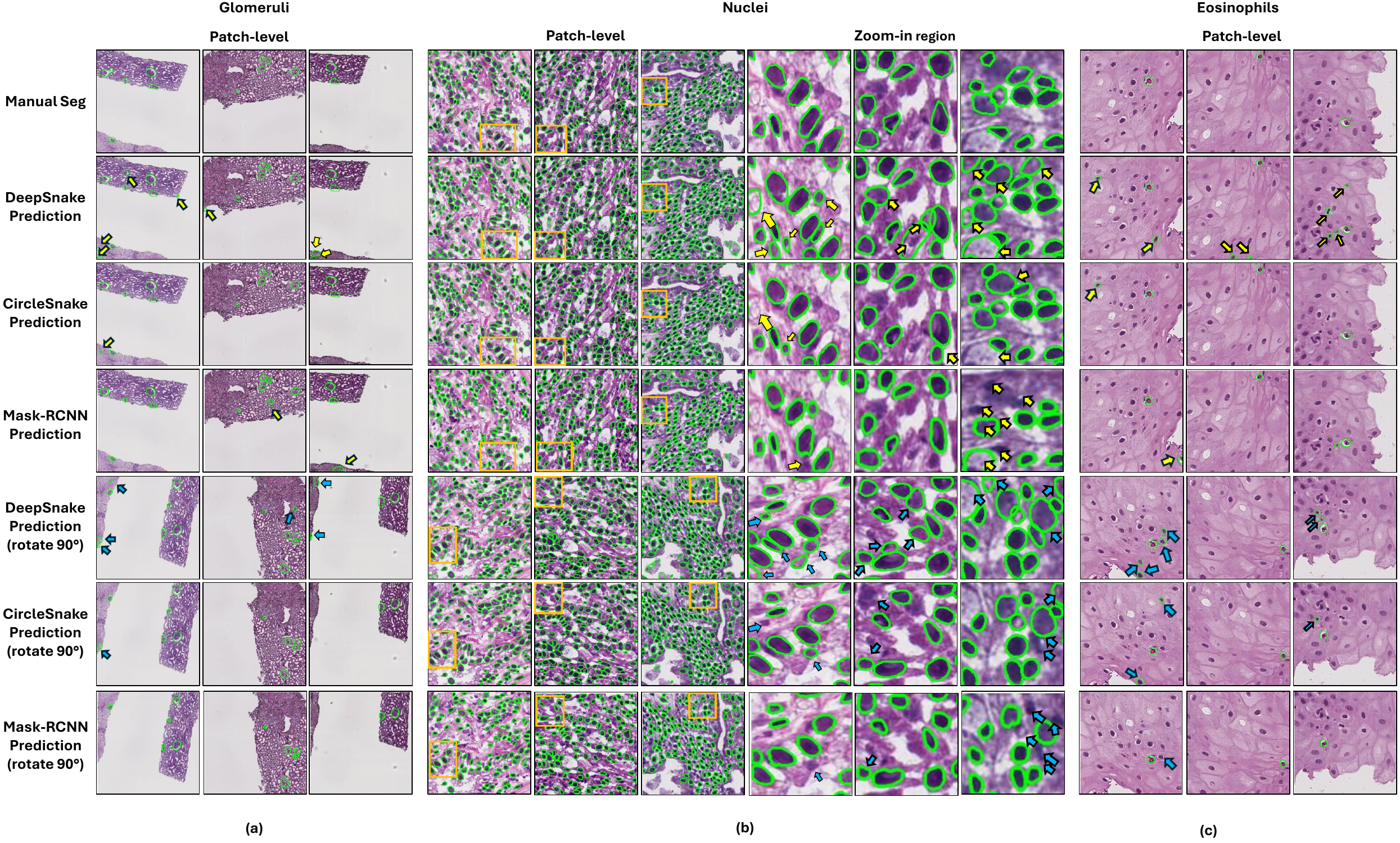}}
\end{center}
\caption{\textbf{Qualitative Comparison.} This figure shows the Qualitative comparison between deepsnake segmentation results and circlesnake segmentation results. Each orange box indicates the location of each selection. Each yellow arrow shows the differences between the manual segmentation results and prediction results . Each blue arrow shows the differences only between the manual segmentation results and the results in rotated 90-degree prediction. (a) is for the Glomeruli dataset, (b) is for the Nuclei dataset, and (c) is for the Eosinophils dataset.}
\label{Fig.4} 
\end{figure*}


\begin{table*}
\caption{Detection and Instance Segmentation Performance on Glomeruli.}
\centering
\begin{tabular}{lcccccc}
 \hline
Methods & Backbone & $AP$ & $AP_{(50)}$ & $AP_{(75)}$ & $AP_{(S)}$ & $AP_{(M)}$\\
 \hline
CenterNet \cite{law2018cornernet}& DLA-34 & 0.547 & 0.872 & 0.643 & 0.432 & 0.640\\
CircleNet \cite{nguyen2022circlesnake} & DLA-34 & 0.570 & 0.857 & 0.691 & 0.466 & 0.647\\
Faster-RCNN \cite{ren2015faster}&ResNet-50 & 0.578 & 0.878 & 0.691 & 0.466 & 0.670\\
Mask-RCNN (Detection) \cite{he2018mask} & ResNet-50 & 0.576 & 0.877 & 0.675 & 0.470 & 0.660\\
Mask-RCNN (Segmentation) \cite{he2018mask} & ResNet-50 & 0.590 & 0.878 & 0.705 & 0.453 & 0.685\\
DeepSnake (Detection) \cite{peng2020deep} &  DLA-34 & 0.527 & 0.881 & 0.609 & 0.424 & 0.610\\
DeepSnake (Segmentation) \cite{peng2020deep} & DLA-34 & 0.548 & 0.877 & 0.669 & 0.411 & 0.642\\
 \hline
CircleSnake (Detection) (Ours) & DLA-34  & 0.603 & 0.877 & 0.734 & \textbf{0.534} & 0.670\\
CircleSnake (Segmentation) (Ours) & DLA-34 & \textbf{0.623} & \textbf{0.894} & \textbf{0.762} & 0.488 & \textbf{0.719}\\
 \hline
\end{tabular}
\label{table1}
\end{table*}

\begin{table*}
\caption{Detection and Instance Segmentation Performance on Nuclei.}
\centering
\begin{tabular}{lcccccc}
 \hline
Methods & Backbone & $AP$ & $AP_{(50)}$ & $AP_{(75)}$ & $AP_{(S)}$ & $AP_{(M)}$\\
 \hline
CenterNet \cite{law2018cornernet} & DLA-34 & 0.404 & 0.818 & 0.342 & 0.440 & 0.290\\
CircleNet \cite{nguyen2022circlesnake} & DLA-34 & 0.461 & 0.843 & 0.462 & 0.473 & 0.332\\
Faster-RCNN \cite{ren2015faster}&ResNet-50 & 0.364 & 0.679 & 0.349 & 0.365 & 0.336\\
Mask-RCNN (Detection) \cite{he2018mask} & ResNet-50 & 0.369 & 0.679 & 0.359 & 0.367 & 0.348\\
Mask-RCNN (Segmentation) \cite{he2018mask} & ResNet-50 & 0.366 & 0.678 & 0.369 & 0.365 & 0.446\\
DeepSnake (Detection) \cite{peng2020deep} &  DLA-34 & 0.401 & 0.827 & 0.315 & 0.408 & 0.345\\
DeepSnake (Segmentation) \cite{peng2020deep} & DLA-34 & 0.425 & 0.827 & 0.394 & 0.425 & \textbf{0.454}\\
 \hline
CircleSnake (Detection) (Ours) & DLA-34  & \textbf{0.485} & \textbf{0.845} & \textbf{0.518} & \textbf{0.495} & 0.217\\
CircleSnake (Segmentation) (Ours) & DLA-34 & 0.436 & 0.837 & 0.419 & 0.436 & 0.418\\
 \hline
\end{tabular}
\label{table2}
\end{table*}

\begin{table*}
\caption{Detection and Instance Segmentation Performance on Eosinophils.}
\centering
\begin{tabular}{lcccccc}
 \hline
Methods & Backbone & $AP$ & $AP_{(50)}$ & $AP_{(75)}$ & $AP_{(S)}$ & $AP_{(M)}$\\
 \hline
CenterNet \cite{law2018cornernet}& DLA-34 & 0.327 & \textbf{0.729} & 0.230 & 0.351 & 0.318\\
CircleNet \cite{nguyen2022circlesnake} & DLA-34 & 0.340 & 0.706 & 0.292 & \textbf{0.368} & 0.332\\
Faster-RCNN \cite{ren2015faster}&ResNet-50 & 0.331 & 0.696 & 0.237 & 0.332 & 0.343\\
Mask-RCNN (Detection) \cite{he2018mask} & ResNet-50 & 0.335 & 0.700 & 0.246 & 0.336 & 0.354\\
Mask-RCNN (Segmentation) \cite{he2018mask} & ResNet-50 & 0.330 & 0.690 & 0.243 & 0.282 & 0.391\\
DeepSnake (Detection) \cite{peng2020deep} &  DLA-34 & 0.300 & 0.720 & 0.149 & 0.301 & 0.323\\
DeepSnake (Segmentation) \cite{peng2020deep} & DLA-34 & 0.305 & 0.712 & 0.169 & 0.255 & 0.360\\
 \hline
CircleSnake (Detection) (Ours) & DLA-34  & \textbf{0.344} & 0.727 & \textbf{0.298} & 0.340 & 0.358\\
CIrcleSnake (Segmentation) (Ours) & DLA-34 & 0.332 & 0.723 & 0.263 & 0.282 & \textbf{0.403}\\
 \hline
\end{tabular}
\label{table3}
\end{table*}

\section{Results}

\begin{figure*}
\begin{center}
\includegraphics[width=0.99\textwidth]{{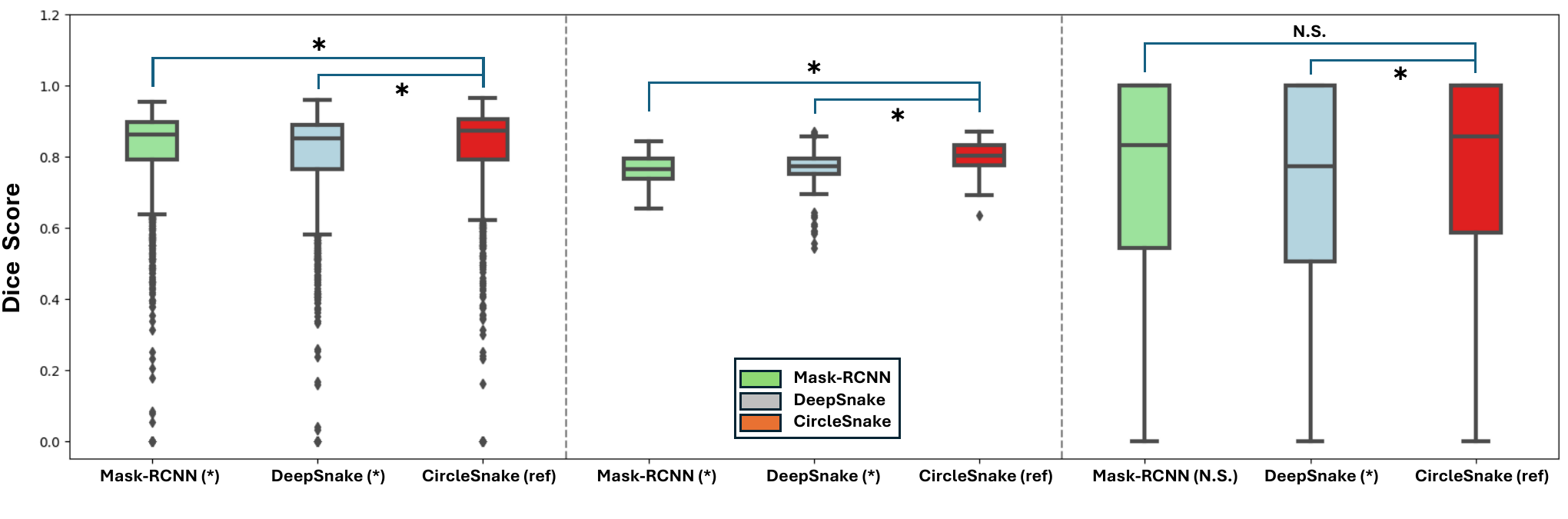}}
\end{center}
\caption{\textbf{Statistical Analysis for Dice Score}The figure shows the boxplots of dice score for Mask-RCNN, DeepSnake, and CircleSnake on Glomeruli, Nuclei, and Eosinophils test datasets. The Wilcoxon signed-rank test is performed with CircleSnake as the reference (``re") method, to compare with other methods. ``*" represents the significant (p $<$ 0.05) differences, while ``N.S.” means the difference is not significant. (a) is for the Glomeruli dataset, (b) is for the Nuclei dataset, and (c) is for the Eosinophils dataset.}
\label{Fig.5} 
\end{figure*}

\begin{figure*}
\begin{center}
\includegraphics[width=0.99\textwidth]{{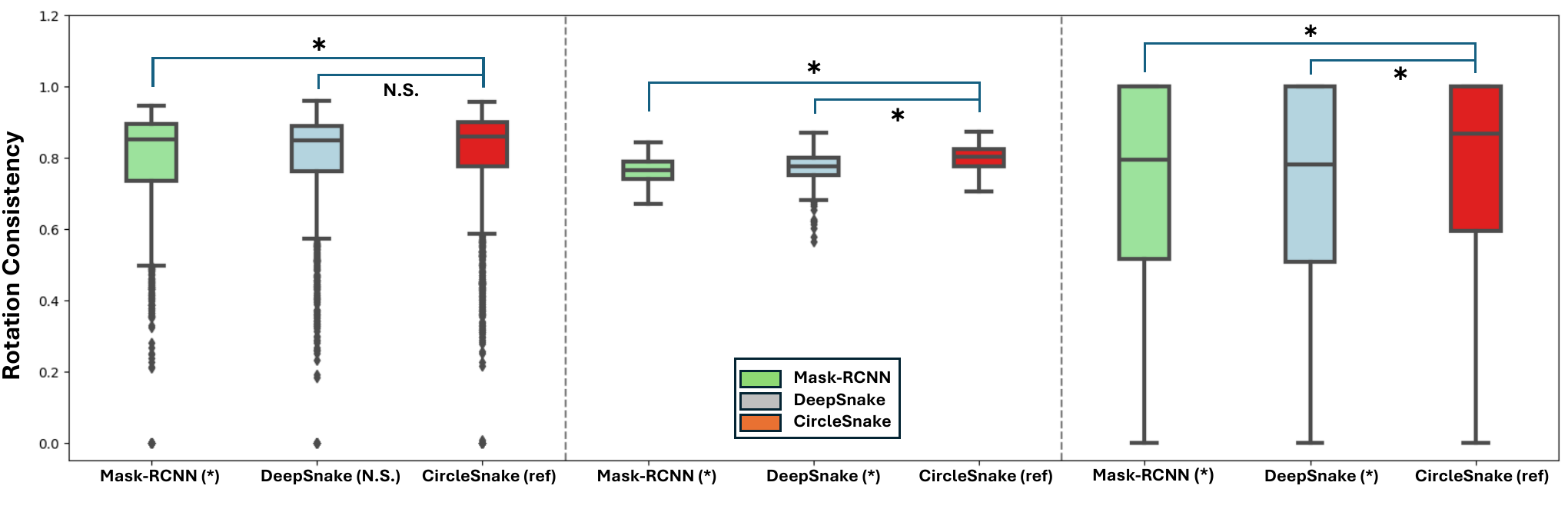}}
\end{center}
\caption{\textbf{Statistic Analysis for Rotation Consistency Score} The figure shows the boxplots of rotation consistency score for Mask-RCNN, DeepSnake, and CircleSnake on Glomeruli, Nuclei, and Eosinophils test datasets. The Wilcoxon signed-rank test is performed with CircleSnake as the reference (``re") method, to compare with other methods. ``*" represents the significant (p $<$ 0.05) differences, while ``N.S.” means the difference is not significant. (a) is for the Glomeruli dataset, (b) is for the Nuclei dataset, and (c) is for the Eosinophils dataset.}
\label{Fig.6} 
\end{figure*}

\begin{table*}
\caption{Semantic Segmentation Results and Rotation Consistency.}
\centering
\begin{tabular}{llccc}
\hline
Dataset & Methods & Backbone & Dice & Rotation Consistency \\
\hline
Glomeruli & Mask-RCNN~\cite{he2018mask} & ResNet & 0.813 & 0.779 \\
Glomeruli & DeepSnake-DLA~\cite{zhou2019objects} & DLA & 0.801 & 0.793 \\
Glomeruli & CircleSnake (Ours) & DLA & \textbf{0.828} & \textbf{0.796} \\ 
\hline
Nuclei & Mask-RCNN~\cite{he2018mask} & ResNet & 0.763 & 0.763 \\
Nuclei & DeepSnake-DLA~\cite{zhou2019objects} & DLA & 0.765 & 0.767 \\
Nuclei & StarDist~\cite{weigert2022} & U-Net & 0.618 & 0.621 \\
Nuclei & CircleSnake (Ours) & DLA & \textbf{0.800} & \textbf{0.799} \\ 
\hline
Eosinophils & Mask-RCNN~\cite{he2018mask} & ResNet & 0.710 & 0.691 \\
Eosinophils & DeepSnake-DLA~\cite{zhou2019objects} & DLA & 0.684 & 0.690 \\
Eosinophils & CircleSnake (Ours) & DLA & \textbf{0.743} & \textbf{0.752} \\ 
\hline
\label{table4}
\end{tabular}
\end{table*}

\begin{table*}
\caption{Semantic Segmentation Results and Rotation Consistency. (training with rotation augmentation)}
\centering
\begin{tabular}{llccc}
\hline
Dataset & Methods & Backbone & Dice & Rotation Consistency \\
\hline
Glomeruli & Mask-RCNN~\cite{he2018mask} & ResNet & 0.829 & 0.839 \\
Glomeruli & DeepSnake-DLA~\cite{zhou2019objects} & DLA & 0.819 & 0.830 \\
Glomeruli & CircleSnake (Ours) & DLA & \textbf{0.834} & \textbf{0.847} \\ 
\hline
Nuclei & Mask-RCNN~\cite{he2018mask} & ResNet & 0.768 & 0.763 \\
Nuclei & DeepSnake-DLA~\cite{zhou2019objects} & DLA & 0.797 & 0.787 \\

Nuclei & CircleSnake (Ours) & DLA & \textbf{0.804} & \textbf{0.805} \\ 
\hline
Eosinophils & Mask-RCNN~\cite{he2018mask} & ResNet & 0.747 & 0.746 \\
Eosinophils & DeepSnake-DLA~\cite{zhou2019objects} & DLA & 0.732 & 0.740 \\
Eosinophils & CircleSnake (Ours) & DLA & \textbf{0.751} & \textbf{0.750} \\ 
\hline
\label{table5}
\end{tabular}
\end{table*}

\subsection{Glomeruli Detection and Segmentation Performance}
As seen in \Tab~\ref{table1}, the proposed CircleSnake method using the DLA as a backbone achieves 0.623 segmentation $AP$, 0.894 segmentation $AP_{(50)}$, 0.762 $AP_{(75)}$, 0.543 detection  $AP_{(S)}$, 0.719 segmentation $AP_{(M)}$. A qualitative comparison between DeepSnake and Circle-Snake can be seen in (\Fig~\ref{Fig.4}) part (a). The arrows are displayed only in areas where the predictions differ from the manual segmentation. Yellow arrows indicate discrepancies in predictions without rotation compared to the manual segmentation, while blue arrows highlight discrepancies in predictions after a 90-degree rotation compared to the manual segmentation.

\subsection{Nuclei Detection and Segmentation Performance}
The dataset from the 2018 Multi-Organ Nuclei Segmentation Challenge~\cite{kumar2017dataset, kumar2019multi} was also applied in the experiment. As seen in \Tab~\ref{table2}, the proposed CircleSnake-DLA method reach 0.485 detection $AP$, 0.845 detection $AP_{(50)}$, 0.518 $AP_{(75)}$, 0.495 detection  $AP_{(S)}$, 0.418 segmentation $AP_{(M)}$. A qualitative comparison between Mask-RCNN, DeepSnake, and CircleSnake can be seen in \Fig~\ref{Fig.4} section (b). 

\subsection{Eosinophils Detection and Segmentation Performance}
In pursuit of confirming the model's broad applicability and generalizability, we decided to introduce the Eosinophils dataset~\cite{liu2023eosinophils} into our study. As indicated in \Tab~\ref{table3}, the performance of the CircleSnake achieve 0.344 detection $AP$, 0.727 detection $AP_{(50)}$, 0.298 $AP_{(75)}$, 0.340 detection  $AP_{(S)}$, 0.403 segmentation $AP_{(M)}$. A qualitative comparison between Mask-RCNN, DeepSnake, and CircleSnake can be seen in \Fig~\ref{Fig.4} part (c).

\subsection{Semantic Segmentation Results and Rotation Consistency}
For the semantic segmentation results and rotation consistency comparison. As shown in \Tab~\ref{table4} and \Fig~\ref{Fig.5} part (a). Our method, CircleSnake can achieve a Dice score of 0.828 with 0.796 rotation consistency score and the Dice score result has a significant difference from the other two methods in the Glomeruli dataset. For the Nuclei dataset \Tab~\ref{table4} and \Fig~\ref{Fig.5} part (b), CircleSnake achieves 0.8 for the Dice score which also has a significant difference from the other two methods and 0.799 in the rotation consistency score. In the Eosinophils dataset \Tab~\ref{table4} and \Fig~\ref{Fig.5} part (c), CircleSnake achieved 0.743 for the Dice score which has a significant difference from the DeepSnake method and 0.752 in the rotation consistency score.

\section{Ablation Study}
As seen in Table \ref{table6}, the StarDist~\cite{weigert2022} U-Net-based segmentation method, which is a widely used approach designed for instance segmentation of star-convex objects such as nuclei, achieves a Dice Score of 0.618 and a Rotation Consistency Score of 0.621 on the Nuclei dataset. StarDist uses a U-Net backbone to predict distances to object boundaries and classifies object centers, making it particularly effective for segmenting objects with distinct shapes. In comparison, our method, CircleSnake, outperforms StarDist on both metrics, demonstrating superior segmentation accuracy and robustness to rotational variations. This highlights the effectiveness of our approach in handling challenging datasets with high object density and complex object boundaries.

\begin{table*}
\caption{Semantic Segmentation Results and Rotation Consistency for StarDist.}
\centering
\begin{tabular}{llccc}
\hline
Dataset & Methods & Backbone & Dice & Rotation Consistency \\
\hline

Nuclei & StarDist~\cite{weigert2022} & U-Net & 0.618 & 0.621 \\
\hline
\label{table6}
\end{tabular}
\end{table*}

\begin{figure}[h]
\begin{center}
\includegraphics[width=0.48\textwidth]{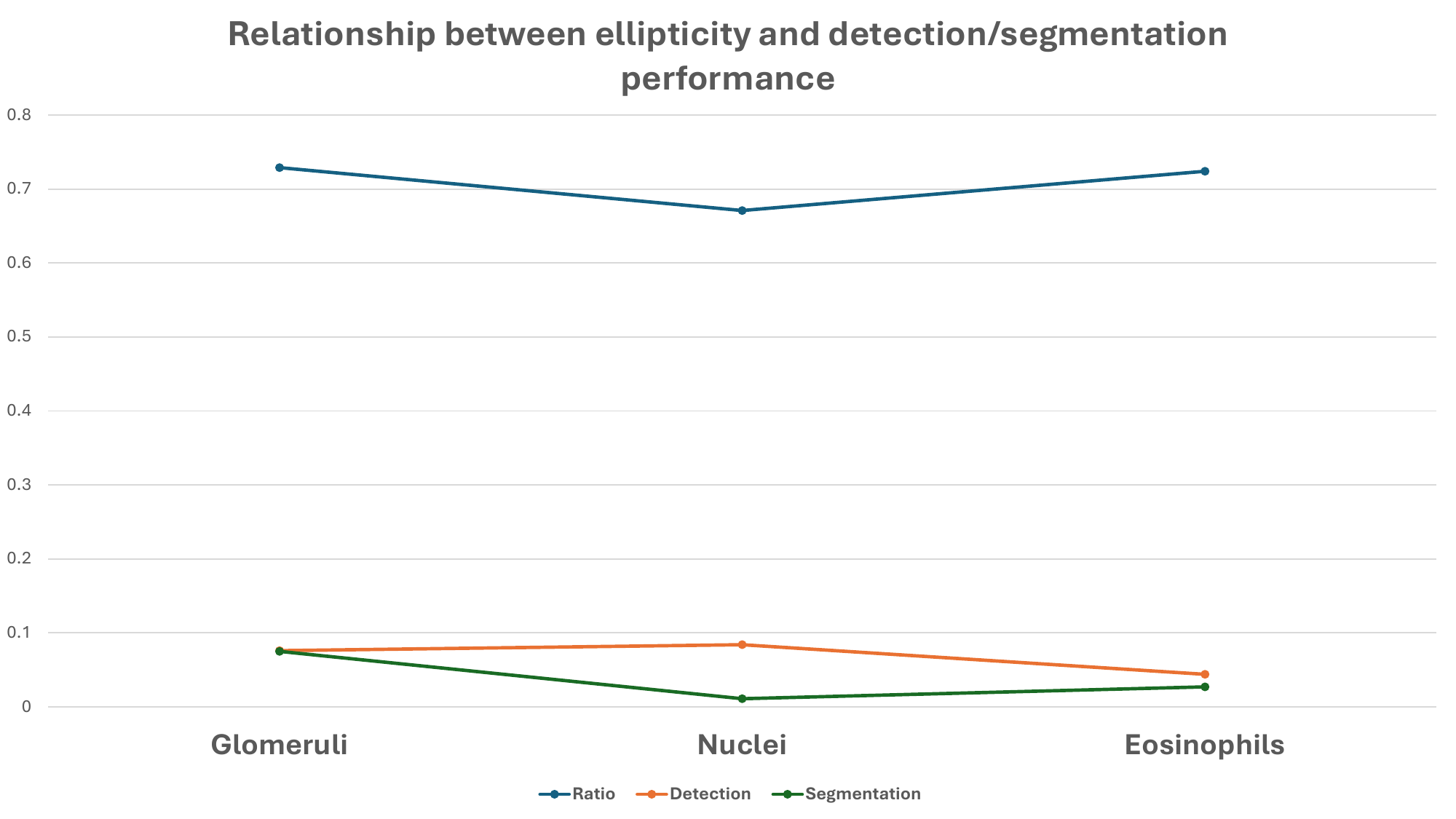}
\end{center}
   \caption{\textbf{Relationship between ellipticity and detection/segmentation performance.} The blue line shows the average ellipticity ratio for all objects on three different datasets. The orange line shows the detection performance on three different datasets. The green line shows the segmentation performance on three different datasets. }
\label{Fig.7}
\end{figure}

\section{Discussion}
In this study, we propose a contour-based method, CircleSnake, optimized for the segmentation of biomedical ball-shaped objects. Instead of using an octagon as the initial contour, CircleSnake directly uses the detection circle as the initial contour which has a simpler structure and is shown to offer superior detection or segmentation performance and rotation consistency. Specifically, CircleSnake outperforms the other two baseline methods in all evaluation metrics in the Glomeruli dataset in \Tab~\ref{table1}. It also has better performance except $AP_{M}$ in the Nuclei dataset in \Tab~\ref{table2} and also better results in $AP$, $AP_{75}$, and $AP_{M}$ in \Tab~\ref{table3}. From \Tab ~\ref{table3}, it can be observed that all models show a noticeable drop in performance on the eosinophil dataset. We believe this is primarily due to the less distinct characteristics of eosinophil cells. As shown in Figure \ref{Fig.4}, glomeruli exhibit the most prominent features, with well-defined boundaries that clearly separate the object from the background. Similarly, nuclei are relatively easier to distinguish due to their strong color contrast with the background. In contrast, eosinophil cells are characterized by intracellular inflammation, often appearing partially red, which makes their features more ambiguous and challenging for all models to detect effectively. This highlights the difficulty of accurately segmenting objects with subtle or less distinct features.


As shown in Tables~\ref{table4} and \ref{table5}, adopting a circular deformation for the segmentation contour leads to better rotation consistency, regardless of whether rotation data augmentation is applied. This improved performance likely stems from the fact that while length and width measurements are sensitive to orientation, radii are inherently more rotation-invariant. To confirm this, we fit the segmentation annotations to ellipses and compute the mean ratio of their minor to major axes (where a ratio closer to 1 indicates a more rounded shape). We then examine how object roundness influences the performance gap in detection and segmentation between CircleSnake and DeepSnake. Our results demonstrate that while CircleSnake’s segmentation performance improves as objects become more circular, no similar trend is observed for detection performance, as illustrated in Figure~\ref{Fig.7}.

Turning to the comparisons in Tables~\ref{table1}, \ref{table2}, and \ref{table3}, we observe that the Nuclei dataset—despite having the highest object density—does not yield the poorest performance. This finding suggests that object density alone does not dictate model success. Instead, COCO AP’s density-aware evaluation ensures that models capable of handling dense object distributions, such as those found in the Nuclei dataset, are fairly assessed for their ability to separate, identify, and accurately segment individual instances. Through this analysis, several factors contributing to errors emerge. For false positives, complex or irregular backgrounds may mimic object features, and overlapping objects in dense scenes can cause over-segmentation of non-existent instances. For false negatives, faint objects or those that are low-contrast relative to the background may be missed, while extreme size variation—particularly very small objects—increases the likelihood of detection failures.

The utilization of a circular contour as the foundational shape for deformation processes may not be ideally suited for objects of varying geometrical forms, such as ellipses or elongated shapes. This assertion is substantiated by the data presented in \Tab~\ref{table2} and \Tab~\ref{table3}, where a discernible discrepancy is observed between the AP of object detection and that of segmentation. The higher AP in detection compared to segmentation might be attributable to the inherent limitations of a circular contour in accommodating the deformation needs of non-circular objects, necessitating a more extensive modification to fit diverse shapes. Furthermore, the current methodology, being predicated on a single-scale feature map, exhibits a marginal reduction in efficacy when dealing with objects of disparate scales, as evidenced by the results in \Tab~\ref{table3}. Consequently, it is posited that transitioning from a single-scale to a multi-scale feature map could potentially enhance the overall performance of the model, providing a more nuanced and adaptable approach to object detection and segmentation.

For the dataset utilization, these three datasets employed in this work are all within the domain of digital pathology, leveraging whole slide imaging. In the future, it would be beneficial to extend the proposed circle represent-ation-based segmentation approach to other fields, such as lung nodule detection (CT slices), liver tumor detection (CT slices), cyst detection (MRI scans), and lesion detection in fundus images. Notably, the circle representation has already been applied in other imaging modalities for object detection, including aerial images~\cite{huang2023semantic} and radiological MR images~\cite{CAMARASA2023102934}, demonstrating its potential versatility across diverse applications.

\section{Conclusion}
In this paper, we propose CircleSnake, a simple contour-based end-to-end deep learning algorithm with simple architecture, that achieves (1) circle detection, (2) circle contour proposal, and (3) circular convolution. The CircleSnake method is optimized for ball-shaped biomedical objects, offering superior glomeruli, nuclei, and eosinophils instance segmentation performance and rotation consistency. The experimental results demonstrate that, in contrast to conventional bounding box and octagonal representations, the reduction of DoF has no adverse impact on the model's accuracy. Furthermore, it effectively sustains detection efficiency across diverse viewing angles. 


\section*{Acknowledgment}
This research was supported by NIH R01DK135597(Huo), DoD HT9425-23-1-0003(HCY), NIH NIDDK DK056942-(ABF). This work was also supported by Vanderbilt Seed Success Grant, Vanderbilt Discovery Grant, and VISE Seed Grant. This project was supported by The Leona M. and Harry B. Helmsley Charitable Trust grants G-1903-03793 and G-2103-05128. This research was also supported by NIH grants R01EB033385, R01DK132338, REB017230, R01MH125931, and NSF 2040462. We extend gratitude to NVIDIA for their support by means of the NVIDIA hardware grant.

%

\section*{Data availability}
The 2018 Multi-Organ Nuclei Segmentation challenge dat-aset can be found at \newline\url{https://monuseg.grand-challenge.org/Data/}. For the Glomeruli dataset and Eosinophils dataset, both are proprietary datasets developed in-house. We currently are unable to share the data due to stringent privacy and confidentiality agreements safeguarding the personal and sensitive information of the individuals involved.

\bibliographystyle{IEEEtran}
\bibliography{sample}
\end{document}